\documentclass{article}

 \usepackage[preprint]{neurips_2026}

\usepackage{listings}
\lstdefinelanguage{JSON}{
  basicstyle=\ttfamily\small,
  breaklines=true,
  showstringspaces=false,
  columns=fullflexible,
  frame=single,
  numbers=left,
  numberstyle=\tiny\color{gray},
  keywordstyle=\color{blue},
  stringstyle=\color{red},
  commentstyle=\color{green}
}
\usepackage{tocloft}
\usepackage{titletoc}
\usepackage{etoolbox}
\usepackage{titlesec}

\usepackage{amsmath,amsfonts,bm}

\def\eqref#1{equation~\ref{#1}}

\def\1{\bm{1}}

\DeclareMathAlphabet{\mathsfit}{\encodingdefault}{\sfdefault}{m}{sl}
\SetMathAlphabet{\mathsfit}{bold}{\encodingdefault}{\sfdefault}{bx}{n}

\usepackage{url}
\usepackage[final,nopatch=footnote]{microtype}
\usepackage[export]{adjustbox}
\usepackage{graphicx}
\usepackage{subfig}
\usepackage{tikz}
\usepackage{bbm}
\usepackage{booktabs} 
\usepackage{enumitem}
\usepackage{xargs}
\usepackage{algorithm}
\usepackage[noend]{algpseudocode}
\usepackage{eqparbox}
\usepackage{multirow}
\usepackage{makecell}
\usepackage{amsmath,amssymb,mathtools}
\usepackage{bigints}
\usepackage{dsfont}
\usepackage{wrapfig}
\usepackage{natbib}
\usepackage{varwidth}
\usepackage[colorlinks, linkcolor=myblue, citecolor=gray]{hyperref}
\usepackage{float}
\usepackage{nicefrac}       
\usepackage{xurl}

\usepackage[most]{tcolorbox}
\usepackage{pifont}
\usepackage[table]{xcolor}

\usepackage{stackengine}
\usepackage{relsize,exscale}
\usepackage{mathrsfs}

\title{Variational Proximal Policy Optimization}

\author{%
  Ousmane Amadou~Dia
  \\
  \texttt{ousamdia@gmail.com} \\
}

\definecolor{mydarkblue}{rgb}{0,0.08,0.45}
\definecolor{mydarkred}{rgb}{0.6,0,0}
\definecolor{myblue}{HTML}{268BD2}
\definecolor{mygreen}{HTML}{658354}

\definecolor{plot_blue}{HTML}{8DC3E5}
\definecolor{plot_green}{HTML}{A4D8A7}
\definecolor{plot_yellow}{HTML}{EFDAA4}
\definecolor{plot_red}{HTML}{DEABB9}

\tcbset{
  mm/prompt/base/.style={
    enhanced, breakable,
    colback=gray!5!white,
    colframe=gray!75!black,
    boxrule=0.6pt, arc=0.8mm,
    left=1.2mm, right=1.2mm, top=1.2mm, bottom=1.2mm,
    before skip=6pt, after skip=6pt,
    fonttitle=\large\bfseries, coltitle=white,
    title={},               
    before upper=\setlength{\parskip}{2pt}\footnotesize
  },
  mm/prompt/light/.style={
    enhanced, breakable,
    colback=gray!4!white,
    colframe=gray!60!black,
    boxrule=0.6pt, arc=0.8mm,
    left=1.2mm, right=1.2mm, top=1.2mm, bottom=1.2mm,
    before skip=6pt, after skip=6pt,
    fonttitle=\bfseries, coltitle=black,
    title={},
    before upper=\setlength{\parskip}{2pt}\small
  }
}

\newtcolorbox{PromptBox}[1][]{mm/prompt/base,#1}
\newtcolorbox{PromptBoxLight}[1][]{mm/prompt/light,#1}

\begin{document}

\maketitle

\begin{abstract}
Reinforcement Learning from Human Feedback via Proximal Policy Optimization often suffers from policy mode collapse, brittle exploration loops, and distribution drift. This paper introduces Variational Proximal Policy Optimization (\(\textsc{VP}_2\textsc{O}\)), a particle-based variational inference framework that maps policy optimization to Stein Variational Gradient Descent within a Mixture-of-Experts architecture. By leveraging functional kernels over localized expert prototypes alongside an expert orthogonalization loss, \(\textsc{VP}_2\textsc{O}\) introduces a geometry-based proximal-control mechanism that can reduce reliance  on fixed clipping or KL schedules. Our results on a 33B/4B sparse Mixture-of-Experts model show several improvements across complex reasoning benchmarks, establishing a \(+\mathbf{179}\) ELO gain on Codeforces and a \(\mathbf{32\%}\) reduction in token count on AIME mathematical reasoning tasks.
\end{abstract}

\section{Introduction}
Reinforcement Learning from Human Feedback (\textsc{RLHF}) has emerged as a powerful paradigm for aligning AI systems with human preferences, enabling applications ranging from dialogue generation, code completion, to mathematical reasoning. However, \textsc{RLHF} pipelines built on Proximal Policy Optimization (\textsc{PPO})~\citep{schulman2017proximalpolicyoptimizationalgorithms} face persistent limitations. First, policies often collapse to a narrow set of high-reward behaviors, sacrificing the diversity needed to capture the full spectrum of human preferences~\citep{casper2023openproblemsfundamentallimitations}. Second, exploration remains inefficient, particularly in sparse or noisy reward landscapes~\citep{wang2024secretsrlhflargelanguage}. Third, policies can overfit to misspecified reward models, leading to reward hacking or distributional drift~\citep{amodei2016concreteproblemsaisafety,10.5555/3618408.3618845}. While recent advances in alignment like \textsc{GRPO}~\citep{shao2024deepseekmathpushinglimitsmathematical} improve stability through gradient penalties, they lack mechanisms for principled exploration and diversity-aware optimization.  

Our approach addresses these challenges by \textit{integrating} Stein Variational Policy Gradient (\text{SVPG})~\citep{liu2017steinvariationalpolicygradient} with \textsc{PPO}, casting policy optimization in \textsc{RLHF} as variational inference over a population of policies. We refer to our approach as \textit{Variational Proximal Policy Optimization} (or \textsc{VP\(_2\)O}). We instantiate \textsc{VP\(_2\)O} within a Mixture-of-Experts (\text{MoE}) architecture, combining it specifically with \textsc{GRPO}, where each expert acts as a distinct policy particle in the variational ensemble.\footnote{Although we combine \textsc{VP\(_2\)O} specifically with \textsc{GRPO}, our framework can be extended to other \textsc{PPO} variants as well.} \textsc{VP\(_2\)O} uses Stein-style transport fields to promote policy diversity and to define geometry-aware
proximal controls, providing an alternative to fixed clipping and KL schedules commonly used in
\textsc{PPO} and variants such as \textsc{GRPO}. 
To further promote structural diversity and prevent expert collapse, we combine the
Stein update with an expert-specialization objective~\citep{guo2026advancingexpertspecializationbetter}, favoring orthogonal expert
representations while reducing routing uniformity and functional overlap within the policy ensemble. 



The key intuition behind \textsc{VP\(_2\)O} is that \textsc{PPO/GRPO}-style \textsc{RLHF} can be viewed through the lens of \textsc{KL}-regularized reward maximization, which induces an optimal policy distribution \(p^*\)~\citep{korbak2022rlkl,rafailov2023direct}. \textsc{VP\(_2\)O} uses this target distribution as the object of a Stein variational update. More intuitively, rather than treating proximal control as a scalar clipping rule, \textsc{VP\(_2\)O} defines a functional geometry over expert prototypes and transports multiple expert particles toward the high-reward regions of \(p^{*}\) while preserving diversity. This innovative framing yields adaptive geometric controls through kernel interactions, \emph{instead of adhoc trust regions enforced via fixed clipping}, particle-based coverage of high-reward behaviors, and gradient smoothing across experts.

We make three key contributions in this study. First, we reinterpret \textsc{KL}-regularized reward maximization in \textsc{RLHF} as minimizing the divergence \( D_{\text{KL}}(\pi_\theta \| p^*)\) using \text{Stein Variational Gradient Descent} (\textsc{SVGD})~\citep{10.5555/3157096.3157362}. \textsc{SVGD} approximates the gradient flow toward the target distribution \( p^* \), replacing \textsc{PPO}’s clipping with kernel-weighted updates that \textit{jointly} optimize reward and diversity. Second, we introduce a joint expert specialization objective—combining an orthogonality loss that encourages independent expert representations with a routing diversification loss that sharpens token-to-expert assignments~\citep{guo2026advancingexpertspecializationbetter}---designed to mitigate expert collapse that standard \textsc{PPO/GRPO} policies suffer from by ensuring each expert within the \text{MoE} develops a functionally distinct behavior. Finally, we unify posterior sampling and policy constraints into a single framework, \textsc{VP\(_2\)O}, enabling policies that are diverse, uncertainty-aware, and aligned with human preferences.

\section{Preliminaries}
Standard \textsc{PPO/GRPO}-style \textsc{RLHF} optimize policies through clipped or regularized objectives that, while effective, face persistent limitations~\citep{ahmadian2024basicsrevisitingreinforcestyle}: (i) they often converge to single modes, sacrificing diversity; (ii) their exploration relies on heuristics (\textit{e.g.} entropy bonuses); and (iii) their trust regions are enforced via ad-hoc constraints (\textit{e.g.} clipping or fixed \(D_{\text{KL}}\) penalties) or heuristics which suffer from high-variance importance sampling~\citep{thomas2015high,mahmood2014weighted}. Token-level importance ratios can become in particular a major source of variance and instability during \textsc{RL} training~\citep{zheng2023clickcontrollabletextgeneration,zheng2025groupsequencepolicyoptimization}: ratios spike under actor-learner drift, destabilizing training and complicating credit assignment across long sequences; clipping becomes a hand-tuned band-aid that distorts gradients and requires per-task tuning, refresh timing turns ad-hoc; and replay gets fragile. For large language models, this materializes as amplified variance, which accumulates over long sequences,  wasted rollouts or over-refreshed actors, and brittle convergence.

As demonstrated below, the \textsc{PPO/GRPO}'s objective can be reformulated as minimizing a \( D_{\text{KL}}(\pi_\theta \| p^*)\) between a target policy \(\pi_{\theta}\) and the optimal policy distribution \( p^* \). This recasting reveals a deeper connection: \textit{\textsc{PPO/GRPO}'s heuristic constraints are approximations of a principled variational problem}, where \textit{trust region constraints} can be replaced with \textit{geometric controls} to stabilize optimization, reduce reliance on \textsc{IS}, and provide data-driven refresh criteria instead of arbitrary refresh schedules.

\subsection{PPO Objective}
\label{sec:sd}
\textsc{PPO}'s surrogate objective (ignoring clipping for now, for brevity) is given by~\citep{schulman2017proximalpolicyoptimizationalgorithms}:
\begin{equation}
    \begin{split}
        \mathcal{J}_{\textsc{PPO}}(\pi_\theta) &= \mathbb{E}_{x\sim\mathcal{D}}\left[\mathbb{E}_{y\sim\pi_{\theta_{\text{old}}}}\left[ \sum_{t=1}^T\kappa_{t}(\theta)\cdot \hat{A}(x, y) 
        \right]\right] \text{ where } \kappa_{t}(\theta) = \frac{\pi_\theta(y_t\mid x, y_{<t})}{\pi_{\theta_{\text{old}}}(y_t\mid x, y_{<t})}.
    \end{split}
    \label{eq:unclipped_ppo}
\end{equation}
Let us define the target distribution---the optimal policy in \textsc{RLHF}---\(p^{*}\) as follows:
\[
  p^{*} = \dfrac{1}{Z(x)}\cdot\pi_{\text{old}}(y \mid x) \exp\Bigg(\dfrac{\hat{A}(x, y)}{\beta}\Bigg) 
\] 
where \( Z(x) = \mathlarger{\int}_y \pi_{\theta_{\text{old}}}(y\mid x) \exp\left( \dfrac{\hat{A}(x, y)}{\beta} \right) d y \) is the partition function (normalization constant).

The forward KL divergence that the unclipped \textsc{PPO} objective (\ref{eq:unclipped_ppo}) implicitly minimizes is:
\begin{equation}
    \begin{gathered}
        D_{\text{KL}}(\pi_\theta \| p^*) = \mathbb{E}_{y \sim \pi_\theta} \left[ \log \pi_\theta(y\mid x) - \log \pi_{\theta_{\text{old}}}(y\mid x) - \dfrac{\hat{A}(x, y)}{\beta} + \log Z(x) \right].
    \end{gathered}
    \label{eq:fwd_kl}
\end{equation}
To see why, let's drop \(\log Z(x)\) from (\ref{eq:fwd_kl}) since it is constant w.r.t \(\theta\) and negate the KL divergence:
\[
 -D_{\text{KL}}(\pi_\theta \| p^*) = \mathbb{E}_{y \sim \pi_\theta} \left[ \dfrac{\hat{A}(x, y)}{\beta} - \log \dfrac{\pi_\theta(y\mid x)}{\pi_{\theta_{\text{old}}}(y\mid x)} \right].
\]
\textsc{PPO/GRPO} approximates the expectation under  \(\pi_{\theta}\) using reweighted samples from \(\pi_{\theta_{\text{old}}}\) as generating sequences from the target policy and training it at the same time is prohibitively costly. 
\begin{equation}
    \begin{gathered}
\mathbb{E}_{y\sim\pi_{\theta}}\left[\dfrac{\hat{A}(x, y)}{\beta}\right] \approx \mathbb{E}_{y\sim\pi_{\theta_{\text{old}}}}\left[\dfrac{\pi_\theta(y\mid x)}{\pi_{\theta_{\text{old}}}(y\mid x)}\cdot \dfrac{\hat{A}(x, y)}{\beta}\right].
    \end{gathered}
    \label{eq:imp_samp}
\end{equation}
This procedure is known as \textit{importance sampling} (\textsc{IS}). The main intuition behind \textsc{IS} is to estimate the expectation of the advantage \(\hat{A}\) under the target policy \(\pi_{\theta}\) by re-weighting the samples drawn from the behavior policy \(\pi_{\text{old}}\). Crucially, this relies on averaging over multiple samples from the behavior distribution \(\pi_{\text{old}}\) for the importance weight to effectively correct for the distributional mismatch. However, \textsc{PPO/GRPO} applies the importance weight \(\kappa_{t}\) at each token position \(t\). Since this weight is based on a single sample \(y_{t}\) from each next-token distribution \(\pi_{\theta_{\text{ref}}}(\cdot\mid x, y_{<t})\), it fails to perform the intended distribution-correction role. Instead, it introduces high-variance noise into the training gradients, which accumulates over long sequences and is exacerbated by the clipping mechanism.

Following (\ref{eq:imp_samp}), the unclipped \textsc{PPO} objective can be recast as a Monte Carlo estimate of \(D_{\text{KL}}(\pi_\theta \| p^*)\). Although it \textit{omits} the \(\log\) term, which acts as a KL penalty, it compensates for its absence through two heuristics: \textit{early stopping}, which prevents \(\pi_{\theta}\) from drifting too far from \(\pi_{\theta_{\text{old}}}\), and \textit{clipping}, which imposes a trust-region-like constraint that approximately bounds \(D_{\text{KL}}(\pi_{\theta}\|\pi_{\theta_{\text{old}}})\), and by extension, \(D_{\text{KL}}(\pi_{\theta}\|p^*)\). Unclipped \textsc{PPO} maximizes the importance-weighted advantage, the first-order term in \(-D_{\text{KL}}(\pi_\theta \| p^*)\), while clipped \textsc{PPO} heuristically approximates the effect of the omitted \(\log\) term.
\begin{equation}
    \begin{gathered}
-D_{\text{KL}}(\pi_\theta \| p^*) \approx \mathbb{E}_{y\sim\pi_{\theta_{\text{old}}}} \left[ \frac{\pi_\theta(y\mid x)}{\pi_{\theta_{\text{old}}}(y\mid x)}\cdot \dfrac{\hat{A}(x, y)}{\beta} - \log \dfrac{\pi_\theta(y\mid x)}{\pi_{\theta_{\text{old}}}(y\mid x)} \right].
    \end{gathered}
    \label{eq:fwd_approx}
\end{equation}
Instead of directly maximizing (\ref{eq:unclipped_ppo}), we apply Stein Variational Gradient Descent (\textsc{SVGD})~\citep{10.5555/3157096.3157362}, a functional gradient method, to derive a gradient-based update for (\ref{eq:fwd_approx}) to optimize \(\pi_{\theta}\). 

\subsection{Stein Discrepancy}
\label{sec:stein_dis}
As before, let \( p^* \) denote the optimal \textsc{RLHF} policy, \( q \) a proposal distribution (\textit{e.g.,} the target policy \( \pi_\theta \)), and \(\mathcal{F}\) a set of test functions \(\phi\colon \mathbb{R}^{d}\to\mathbb{R}^d\). The \textit{Stein's discrepancy} of \(q\) with respect to \(p^*\) is:
\[
    \mathbb{SD}(q\| p^*) = \sup_{\boldsymbol{\phi}\in\mathcal{F}} \mathbb{E}_{\theta \sim q}[\mathcal{A}_{p^*} \boldsymbol{\phi}(\theta)],
\] 
where \( \mathcal{A}_{p^*} \boldsymbol{\phi}(\theta)= \nabla_{\theta} \log p^*(\theta) \boldsymbol{\phi}(\theta) + \nabla_{\theta} \boldsymbol{\phi}(\theta) \) is the Stein operator.
When \( q = p^* \), the discrepancy is equal to zero. Otherwise, it upper-bounds \( D_{\text{KL}}(q \| p^*) \) under smoothness assumptions. The Stein discrepancy \(\mathbb{SD}(q\| p^*)\) measures how well \( q \) satisfies the \textit{Stein identity} for \( p^* \) under \(\mathcal{F}\):
      \[
      \mathbb{E}_{\theta \sim q}[\mathcal{A}_{p^*} \boldsymbol{\phi}(\theta)] = 0 \quad \forall \boldsymbol{\phi}\in\mathcal{F}.
      \]
To make the optimization tractable, \(\boldsymbol{\phi}\) is restricted to a unit ball in some reproducing kernel Hilbert space \(\mathcal{H}^d\) induced by a positive-definite kernel \(k(x, x')\). By the \textit{representer theorem}, the solution is:
\[
    \boldsymbol{\phi}^*_{q,p^*} (\cdot) = \mathbb{E}_{\theta\sim q}\left[\mathcal{K}(\theta,\cdot)\nabla_{\theta}\log p^*(\theta) + \nabla_{\theta} \mathcal{K}(\theta,\cdot)\right].
\]
To approximate the direction \(\boldsymbol{\phi}^*_{q,p^*}(\cdot)\), a set of \(N\) particles \(\{\theta_i^0\}_{i=1}^N\) from an initial distribution $q_0$ are maintained. Then, the particles are updated using an empirical version of \(\boldsymbol{\phi}^*_{q,p^*}\) in which \(\mathbb{E}_{\theta\sim q_{\ell}}[\cdot]\) is approximated by the mean of the particles at the $\ell$-iteration (see Equation~\ref{eq:sko_grad}). This procedure, called \textit{Stein Variational Gradient Descent} (\textsc{SVGD})~\citep{10.5555/3157096.3157362}, progressively evolves the particles to match the distribution \(p^*\), effectively providing a sampling method for $p^*$. Since \textsc{SVGD} is independent of the distribution \(q_0\), it confers lots of flexibility in how to initialize each particle \(\theta_i\).
\begin{equation}
    \begin{gathered}
        \theta_i^{\ell+1}\leftarrow \theta_i^{\ell} + \eta_{\ell}\boldsymbol{\phi}^*(\theta_i^{\ell}) \text{ with } \boldsymbol{\phi}_{q,p^*}^*(\theta_i^{\ell}) = \dfrac{1}{N}\sum_{j=1}^N \Big[\underbrace{\mathcal{K}(\theta_j^{\ell}, \theta_i^{\ell}) \nabla_{\theta_j} \log p^*(\theta_j^{\ell})}_{\text{Driving Force}} + \underbrace{\nabla_{\theta_j}\mathcal{K}(\theta_j^{\ell}, \theta_i^{\ell})}_{\text{Repulsive Force}} \Big].
    \end{gathered}
    \label{eq:sko_grad}
\end{equation}
SVGD relies on two forces to approximate the target distribution \(p^{*}(\theta)\): a \textit{driving force} which moves particles toward high-likelihood regions, and a \textit{repulsive force} which prevents particle collapse. Both forces may, however, experience weaknesses depending on the choice of the kernel function \(\mathcal{K}(\cdot;\cdot)\).
By casting \textsc{PPO/GRPO}'s objective as variational inference, we approximate \(D_{\text{KL}}(p^{*}\| \pi_{\theta})\) via an ensemble of \textit{particles} \(\{\pi_{\theta_i}\}_{i=1}^N\) that collectively approximate \(p^{*}\)'s entire support using \textsc{SVGD}. For \textsc{RLHF}, the model architecture that lends itself the most to this  optimization is the \textit{mixture} \textit{of experts}.

\subsection{Mixture of Experts}
To realize the variational inference framework described in Section~\ref{sec:stein_dis}, we require a parameterized distribution that can maintain multiple ``modes" or particles of behavior. We leverage the Sparsely-Gated Mixture of Experts (\text{MoE}) architecture \citep{2015arXiv150302531H,shazeer2017outrageouslylargeneuralnetworks}, reinterpreting its modular structure as a discrete ensemble of policy particles. In a standard Transformer-based \text{MoE}, the dense Feed-Forward Network (\textsc{FFN}) layers are replaced by \(N\) independent expert blocks. For any given input token \(x\), a trainable router \(\phi(\cdot)\) selects a sparse subset of \(K\) experts to process the information. While \text{MoE} is traditionally used for computational efficiency---increasing model capacity without a proportional increase in \textsc{FLOP}s---we utilize it here for its functional modularity.

\textbf{Expert Specialization and Particle Identity.} Each expert \(i\) within an \text{MoE} layer acts as a functional ``particle" in the Stein discrepancy framework. Rather than viewing the \text{MoE} as a single monolithic model, we treat the individual experts as distinct policy components \(\pi_{\theta_i}\) that collectively approximate the optimal policy distribution \(p^*\). The \text{MoE} architecture provides two critical properties for our approach: (i) \textit{conditional computation} and (ii) \textit{structural diversity}. For (i), by activating only a subset of Top-\(K\) experts, the model naturally creates high-density regions of the policy space, which we can then regularize using Stein forces. For (ii), the separation of experts allows us to apply the repulsive force (\ref{eq:sko_grad}) directly to the expert parameters or their output projections, preventing the ``mode collapse" common in standard \textsc{RLHF} where the model converges to a single, narrow set of high-reward tokens.

\textbf{Routing as Variational Assignment.} The \text{MoE} gating function \(\omega(x) = \text{softmax}(\phi(x))\) serves as a local assignment mechanism. In our formulation, the router's role is not just load balancing, but also distributional coverage. By analyzing the co-activation patterns of these experts, we can define a kernel \(\mathcal{K}\) that measures the similarity between policy particles in the functional space \(\mathcal{H}^d\). This allows the Stein update to move ``similar" experts toward high-reward regions (via attraction) while pushing ``redundant" experts toward unexplored areas of the preference landscape (via repulsion).

One of the primary benefits of \text{MoE} architectures is their efficiency~\citep{shazeer2017outrageouslylargeneuralnetworks,DBLP:journals/corr/abs-2101-03961,6797059}. Although, the number of experts \(N\) in an \text{MoE} layer can be relatively large, only a subset of the experts are selected to process each token~\citep{DBLP:journals/corr/abs-2101-03961}. This means that only a small portion of the \text{MoE} model parameters---\textit{given by the experts selected at each layer}---are active when processing a given token. Precisely, given a token representation \(x\), each \text{MoE} layer maintains a routing structure \(\phi(\cdot)\) to sparsely select the experts to activate. Such a structure can take one of two forms: a \textit{soft-routing} or \textit{hard-routing}. Formally, soft-routing applies a linear transformation to the token vector \(x\), forming a vector of size \(N\), the number of experts in the layer, followed by a softmax to induce a probability distribution over the set of all experts. Formally: 
\[s(x) = \omega(x) = \texttt{softmax}(\phi(x)) \in [0, 1]^N.\]  
Hard-routing, on the other hand, maintains a binary vector \(\in\{0, 1\}^N\), where the indices of the activated experts are set to \(1\), while the remaining indices are set to \(0\):
\(
s_i(x)=\mathbb{I}\{i\in \text{Top-}K(\omega(x))\}
\). We refer to the selected experts as \textit{co-activated experts} and assign a \textit{co-activation score} \(\boldsymbol{O}_{ij}\in\mathbb{R}\) to any pair of experts \((i, j)_{1\leq i, j \leq N}\) to create a co-activation score matrix \(\boldsymbol{O}\in\mathbb{R}^{N\times N}\) that is defined by:   
\begin{equation}
\begin{gathered}
\boldsymbol{O} \;\triangleq\; \mathbb{E}_{x\sim \mathcal{B}}\Big[s(x)\,s(x)^\top\Big], \quad \boldsymbol{O}_{ij}=\mathbb{E}_{x\sim \mathcal{B}}\Big[s_i(x)s_j(x)\Big]. 
\end{gathered}
\end{equation}
\section{Stein Policy Update}
\subsection{Functional Kernel}
Let \({\Theta} = \{\theta_i\}_{i=1}^N \) denote an \text{MoE} layer  composed of \(N\) \textit{specialized experts} and a router structure \(\phi\). In \textsc{RLHF}, each expert \(i\) induces a policy \(\pi_{\theta_i}\), parameterized by \(\theta_i\), and the mixture policy \(\pi_{\text{mix}}\) is:
\[
\pi_{\text{mix}}(y \mid x) = \sum_{i=1}^N \omega_i(x)\pi_{\theta_i}(y\mid x),\;\; \omega(x) = \texttt{softmax}(\phi(x)) \in [0, 1]^N.
\]
We optimize \(\pi_{\text{mix}}\) toward the optimal policy distribution \(p^{*}\) using \textsc{SVGD} as the driving flow while keeping \textsc{PPO/GRPO}'s advantage estimation approach. To that end, let \(\mathcal{E}(x)\subseteq{1,\dots,N}\) denote the \textsc{Top-}\(K\) experts selected for the token \(x\), and let \(\mathcal{T(B)}=\bigcup_{x\in B} \mathcal{E}(x)\) denote the union of selected experts over a batch of prompts \(\mathcal{B}\). Let us define the expert-level attraction gradient as follows: 
\begin{equation}
\begin{gathered}
g_i \;\triangleq\; \nabla_{\theta_i}\log p^*(\theta_i) = \mathbb{E}_{x}\Bigg\{\mathbb{E}_{y\sim \pi_{\theta_i}(\cdot\mid x)} \Bigg[\dfrac{\hat A(x,y)}{\beta }\,\nabla_{\theta_i}\log \pi_{\theta_i}(y\mid x)\Bigg]\Bigg\}.
\end{gathered}
\end{equation}
We update only the active experts \(i\in \mathcal{T(B)}\) with an \textsc{SVGD}-style transport field \(\varphi(\theta_i)\) as per below:
\[
\theta_i \leftarrow \theta_i + \eta \,\varphi(\theta_i), \quad i\in \mathcal{T(B)}.
\]
The Stein transport field \(\varphi(\theta_i)\) combines an attraction term---\textit{kernel-weighted averaging of gradients}---and a repulsion term---\textit{kernel-gradient term}---with a routing structure \(\phi(\cdot)\) that determines which experts interact. The kernel acts in function space over a union \textsc{Top-}\(K\) vocabulary for efficiency. In practice, we approximate \(\varphi(\theta_i)\) with a prototype-based output-space kernel since parameter-space kernels are brittle in high dimensions. Furthermore, combining \textsc{SVGD’}s repulsive force with \text{MoE}’s specialization introduces two contentious goals:  (i) the routing strategy requires certain experts to be structurally similar (\textit{i.e.,} close in parameter space) in order to handle highly correlated inputs, and (ii) the \textsc{SVGD} repulsive force, which will actively try to push the experts apart. We address this contention by introducing the notion of \textit{expert directionality} to compute the kernel \(\mathcal{K}\). We maintain an output-space prototype \(\mathbf{p}_i\in\mathbb{R}^{D_{\text{out}}}\) with unit-norm for each expert \(i\). This kernel directly measures behavioral/directional similarity, leveraging the angular distances between the experts' prototypes.

Let \(\mathbf{W}_i\in\mathbb{R}^{D_{\text{in}}\times D_{\text{out}}}\) denote expert \(i\)’s output projection matrix (a distinguished subset of parameters within \(\theta_i\) used only for prototype computation). We update \(\mathbf{p}_i\) via n-step power iteration on \(\mathbf{W}_i^\top \mathbf{W}_i\):
\[
\mathbf{p}_{i,n} \leftarrow \dfrac{\mathbf{W}_i^\top \mathbf{W}_i\,\mathbf{p}_{i,n-1}}{\|\mathbf{W}_i^\top \mathbf{W}_i\,\mathbf{p}_{i,n-1}\|_2}, \text{ where } \|\mathbf{p}_{i,n}\|_2=1 \text{ and } \mathbf{p}_{i,0}=\dfrac{\tilde{\mathbf{p}}_{i,0}}{\|\tilde{\mathbf{p}}_{i,0}\|_2} \text{ with } \tilde{\mathbf{p}}_{i,0}\sim\mathcal{N}(0,I). 
\]
The resulting \(\mathbf{p}_i\) (dropping the \(n\) index) is the principal eigenvector of \(\mathbf{W}_i^\top \mathbf{W}_i\), \textit{i.e.,} the top right singular direction of \(\mathbf{W}_i\). It captures expert \(i\)’s dominant output-space direction. We define \(\mathcal{K}\) as:
\begin{equation}
\begin{gathered}
\mathcal{K}_{ij}\ = \dfrac{1}{\tau}\cdot\dfrac{\langle \mathbf{p}_i, \mathbf{p}_j\rangle}{\|\mathbf{p}_i\|\|\mathbf{p}_j\|}\;\;\; \|\mathbf{p}_i\|_2 = \|\mathbf{p}_j\|_2 = 1, \quad \text{ for } 1 \leq i, j \leq N.
\end{gathered}
\end{equation}
To control the scale of interactions as the number of experts \(N\) varies, we introduce a temperature \(\tau = T_0/N\), where \(T_0\) denotes a baseline temperature value. The kernel \(\mathcal{K}\) induces a smooth notion of similarity between experts based on how aligned their dominant output-space directions are. Importantly, \(\mathcal{K}\) depends implicitly on the expert parameters through the mapping \(\mathbf{W}_i\mapsto \mathbf{p}_i\).

\subsection{Stein Policy Update}
Instead of repelling the active experts against each other, which can destabilize local updates, we decouple the Stein transport field into an \textit{attraction term} from co-active neighbors and a \textit{repulsion term} from inactive neighbors. Formally, we introduce two routing-structure masks that we define as
\begin{itemize}
    \item \(M^{\text{att}}\in\{0,1\}^{N\times N}\): co-active neighbors \((e.g., \text{Top-}K\) by \(\boldsymbol{O}_{ij}\) per row, or \(\boldsymbol{O}_{ij}\ge \delta_{\text{att}})\).
    \item \(M^{\text{rep}}\in\{0,1\}^{N\times N}\): inactive/rarely-coactive neighbors \((e.g., \boldsymbol{O}_{ij}\le \delta_{\text{rep}})\).
\end{itemize}
The attraction term encourages experts that are co-activated by \(\phi(\cdot)\) to share information. For expert \(i\), the attraction term is defined as: \(\varphi_{\text{attr}}(\theta_i)=\sum_{j=1}^N \alpha_{ij}\,\mathcal{K}_{ij}\,g_j\), where the weights \(\alpha_{ij}\) are defined as:
\begin{equation}
\begin{gathered}
\alpha_{ij}=M^{\text{att}}_{ij}\Big/\max\Big(1,\sum_k M^{\text{att}}_{ik}\big), \quad \beta_{ij}=M^{\text{rep}}_{ij}\Big/\max\Big(1,\sum_k M^{\text{rep}}_{ik}\Big). 
\end{gathered}
\end{equation}
The repulsion term plays the complementary role of discouraging collapse among experts that are not jointly active. Since \(\mathcal{K}_{ij}\) depends on expert parameters through \(\mathbf{W}_i\mapsto \mathbf{p}_i\), its gradient is non-zero. We formalize the repulsion term using the distinguished matrix \(\mathbf{W}_i\subset\theta_i\): \(\varphi_{\text{rep}}(\theta_i)=\sum_{j=1}^N \beta_{ij}\,\nabla_{\mathbf{W}_i}\mathcal{K}_{ij},\) with the weights \(\beta_{ij}\) defined as above.
This term pushes expert \(i\) away from inactive experts in the induced kernel geometry, encouraging specialization and diversity in the output-space directions. Combining the attraction and repulsion terms produces the transformed gradient applied to expert \(i\):
\begin{equation}
\begin{gathered}
\tilde g_i = \sum_{j=1}^N \alpha_{ij}\mathcal{K}_{ij}g_j\, + \, \lambda_{\text{rep}} \sum_{j=1}^N \beta_{ij}\nabla_{\mathbf{W}_i}\mathcal{K}_{ij}, \text{ with } \nabla_{\mathbf{W}_i}\mathcal{K}_{ij} = \dfrac{1}{\tau} \nabla_{\mathbf{W}_i}\langle \mathbf{p}_i,\mathbf{p}_j\rangle,
\end{gathered}
\end{equation}
where \(\lambda_{\text{rep}}\) controls the strength of the repulsive interaction. This formulation mirrors the classical \textsc{SVGD} update in a few ways: (i) the kernel-weighted gradient averaging term corresponds to attraction toward high-density regions, (ii) the kernel-gradient term provides repulsion that prevents collapse, and (3) the coactive/inactive masks tailor these forces to the \text{MoE} routing structure. This yields an \textsc{SVGD}-style decomposition into \textit{kernel-weighted attraction} among co-active experts and \textit{kernel-gradient repulsion} against inactive neighbors, where routing determines the interaction.
\[
\varphi(\theta_i)=\varphi_{\text{attr}}(\theta_i)+\lambda_{\text{rep}}\varphi_{\text{rep}}(\theta_i), \quad \theta_i \leftarrow \theta_i + \eta\,\varphi(\theta_i). 
\]
As neither soft-routing nor hard-routing explicitly encourages a balanced selection of experts, an \text{MoE} layer is likely to converge to a state where it always produces large weights for the same few experts instead of fully and uniformly utilizing its expert layers. To encourage a balanced selection of experts and prevent experts from starving, auxiliary load balancing losses are generally introduced to further improve training stability. However, such losses often lead to expert overlap and overly uniform routing, which hinders expert specialization and degrades performance~\citep{guo2026advancingexpertspecializationbetter}.

Based on the observation above, we adopt the following design, inspired by~\citep{guo2026advancingexpertspecializationbetter}, to further mitigate expert overlap and routing uniformity. We introduce an orthogonalization objective that encourages independent expert representations. Specifically, we design the orthogonality loss as:
\[
\mathcal{L}_o = \sum_{i=1}^N\sum_{j=1}^N\sum_{\substack{k=1\\ k\neq i}}\Bigg\|\dfrac{\langle\tilde{x}_{ij}, \tilde{x}_{ik}\rangle}{\langle\tilde{x}_{ik}, \tilde{x}_{ik}\rangle + \epsilon}\Bigg\|^2,\quad \tilde{x}_{ij} = x_{ij}\cdot \theta_{j}\cdot \mathbb{I}_{\{o_{ij} > 0\}},
\]
where \(\langle\cdot\rangle\) denotes the inner product between two vectors, \(\mathbb{I}_{\{o_{ij}> 0\}}\) is an indicator function that
evaluates to 1 when \(o_{ij} > 0\) and \(0\) otherwise, and \(\tilde{x}_{ij}\) represents the output of expert \(j\) for token \(x_i\) after the \(\text{Top-}K\) routing selection. \(\mathcal{L}_o\) reduces the overlap between different expert outputs within the same \(\text{Top-}K\) group by minimizing their projections onto each other. This encourages experts to develop more distinct representations, promoting specialization in processing different token types.

\subsection{Geometric Trust Regions}
We replace \textsc{PPO/GRPO}'s clipping constraints with two geometric controls: a \textit{budget anchored to the actors’ policy} in the low-dimensional prototype space and an \textit{on-policy behavior budget}. Such controls encourage better training stability than the ad-hoc  heuristic clipping or fixed KL penalties.

\textbf{Anchor Prototype Budget.} We control the step size in a low-dimensional, router-aware prototype space by anchoring the learner’s updates to the policy snapshot \(\mathbf{p}^{\text{anc}}_i\) currently deployed on the actors. This provides a data-driven, geometric proximal-control signal in prototype space. It complements the likelihood-ratio and \textsc{ESS} diagnostics below by limiting how far the learner’s router-weighted expert prototypes move from the actor snapshot, reducing reliance on fixed clipping thresholds as the primary mechanism for controlling policy drift. The prototypes are updated after each optimizer step, or every fixed number of steps. The anchor snapshots are taken whenever the actors are refitted:
\[
\mathbf{p}_i^{\text{anc}} \coloneqq \mathbf{p}_{i,t},\quad \|\mathbf{p}_{i,t}\|_2=1.
\]
The learner and anchor policies are compared via their prototype barycenters over the rollout batch contexts \(\mathcal{B}\). The actors log their Top-\(K\) gates \(\omega_i^{\text{act}}(x)\), while the learner computes its gates \(\omega_i^{(t)}(x)\): 
\[
\mathbf{P}^{\text{anc}}(x) = \sum_{i=1}^N \omega_i^{\text{act}}(x)\, \mathbf{p}_{i}^{\text{anc}},\quad\mathbf{P}^{(t)}(x)=\sum_{i} \omega_i^{(t)}(x)\,\mathbf{p}_{i,t}.
\]

The primary trust region is defined by the average Euclidean distance between the learner’s current barycenter and the anchor’s barycenter over the batch \(\mathcal{B}\). The step is accepted if and only if this deviation is below a certain threshold value \(\kappa_{\text{proto}}\):
\[
\bar\Delta\mathbf{P}^{(t)} = \dfrac{1}{\mathcal{B}}\sum\limits_{x\in\mathcal{B}}\Big\|\mathbf{P}^{(t)}(x) - \mathbf{P}^{\text{anc}}(x)\Big\|_2\leq \kappa_{\text{proto.}}
\]

\textbf{On-Policy Behavior Budget.} We maintain an explicit on-policy behavior budget between logged behavior probabilities \(\pi_{\text{mix}}^{\text{old}}\) and the current mixture \(\pi_{\text{mix}}\). Rather than refitting after every \(m\geq 1\) learner step(s), we use event-driven synchronization based on measurable drift budgets. The logged likelihood ratios and ESS estimates serve as diagnostics for actor-learner drift: when the drift exceeds \(\epsilon\) or ESS falls below \(\tau_{\text{ess}}\), we refresh the actors. This reduces reliance on clipped importance-weighted updates while keeping the learner close to the behavior policy that is used to generate rollouts.
\[
\mathbb{E}_{\text{tokens}}\Big[ D_{\text{KL}}\big(\pi_{\text{mix}^{\text{old}}}\ \Vert\ \pi_{\mathrm{mix}}\big) \Big]\ \le\ \varepsilon \hspace{1em} \text{ and }\hspace{1em} \textsc{ESS} = \dfrac{\left(\sum_t w_t\right)^2}{\sum_t w_t^2}\geq \tau_{\text{ess}},
\]
where:
\[
\pi_{\text{mix}}(y\mid x) = \sum_{i=1}^N \omega_i(x)\cdot\pi_{\theta_i}(y\mid x) =  \sum_{k=1}^K \omega_k\cdot \pi_{\theta_k}(y\mid x), \hspace{1em} r_t=\dfrac{\pi_{\mathrm{mix}}(y_t\mid x, y_{< t})}{\pi_{\text{mix}^{\text{old}}}(y_t\mid x, y_{<t})},\quad w_t(\alpha)=r_t^\alpha.
\]
\textsc{ESS}, or \emph{effective sample size}, measures how many samples from an older behavior policy \(\mu\) are effectively equivalent to independent on-policy samples. We compute \textsc{ESS} from the tempered likelihood ratios \(w_t(\alpha)=r_t^\alpha\), where \(\alpha\) controls the sensitivity of the diagnostic to actor-learner drift. If the average drift exceeds \(\epsilon\) or the resulting ESS falls below \(\tau_{\text{ess}}\), we schedule an actor refresh to bring the rollout policy back in sync with the learner; that is to stay on-policy:
\begin{equation}
\begin{gathered}
\mathbb{E}_{\text{tokens}}\Big[ D_{\text{KL}}\big(\mu\ \Vert\ \pi_{\mathrm{mix}}\big) \Big]\ >\ \varepsilon \hspace{1em} \text{ or }\hspace{1em} \text{ESS} = \dfrac{\left(\sum_t w_t\right)^2}{\sum_t w_t^2} <\tau_{\text{ess}}.
\end{gathered}
\end{equation}

\section{Experiments}
\label{sec:experiments}
 
\subsection{Experimental Setup}
 
\textbf{Model.}
We evaluate \textsc{VP\(_2\)O} on a 33B/4B-parameter sparse \text{MoE} model 
initialized from a pre-trained checkpoint with \(N=20\) experts per
feed-forward layer and Top-\(K\) routing.  The same architecture and
initialization are shared with the baseline model to isolate the effect of
the training objective.
 
\textbf{Training details.}
Advantages are computed via group-relative policy
optimization~\citep{shao2024deepseekmathpushinglimitsmathematical}.
Optimization uses AdamW~\citep{DBLP:journals/corr/abs-1711-05101}
(\(\beta_1=0.9,\;\beta_2=0.95\)) with a cosine learning-rate schedule.
The Stein kernel temperature is set to \(\tau=T_0/N\) and the repulsion
coefficient to \(\lambda_{\text{rep}}=0.1\).  The orthogonality-loss
weight follows~\citet{guo2026advancingexpertspecializationbetter}'s experimental settings with
\(\beta=0.01\).  We train at two generation lengths (8K and 16K tokens)
to assess sensitivity to context.  Checkpoints are saved at regular
step (after every 80 training steps) intervals throughout training.
 
\textbf{Baseline.}
Our primary comparison is a \textsc{GRPO}-trained \text{MoE} model
trained under identical compute, data, and architectural conditions.
This is the strongest known \textsc{PPO}-variant for our setting.
 
\textbf{Benchmarks.}
We evaluate across the followong five categories:
\begin{itemize}[leftmargin=*]
  \item\textit{Mathematical reasoning:} AIME 2024~\citep{aime24} and AIME
        2025~\citep{aime25}, competition-level problem solving.
        
  \item 
  \textit{Scientific reasoning:} GPQA
        Diamond~\citep{rein2023gpqa},
        graduate-level science QA.
        
  \item 
  \textit{Knowledge:} MMLU~\citep{wang2024mmlu},
        a multi-task language understanding benchmark.
        
  \item 
  \textit{Code generation:} Codeforces ELO and Pass@1, measuring
        competitive-programming ability.
        
  \item 
  \textit{Instruction following:}
        IFBench~\citep{pyatkin2025generalizing} and IFEval~\citep{zhou2023instructionfollowingevaluationlargelanguage},
        evaluated under both loose and strict protocols.
\end{itemize}
Metrics are averaged across all evaluation checkpoints.
 
\subsection{Main Results}
\label{sec:main_results}
 
Table~\ref{tab:main_results} reports the average score of each method
across training checkpoints and generation lenghts. \textsc{VP\(_2\)O} delivers consistent
gains across all benchmark categories, with improvements that generally
\emph{increase} from 8K to 16K, indicating that the variational
framework scales with generation length.
\begin{itemize}[leftmargin=*]
\item\textbf{Mathematical reasoning.} 
\textsc{VP\(_2\)O} improves AIME 2024 and AIME 2025 by \(+\mathbf{2.6}\) and
\(+\mathbf{2.8}\) percentage points (pp) under 8K. In the 16K setting, AIME 2024
converges around \(2{,}000\) steps earlier than the baseline while
retaining a \(+\mathbf{1.6}\) pp lead on average, consistent with the Stein
driving force pooling gradient information across expert particles. While the gains seem modest,
the early convergence shows that \textsc{VP\(_2\)O} reaches strong AIME 2024 performance earlier than the
baseline.

\item\textbf{Scientific and knowledge reasoning.}
GPQA Diamond shows near-parity at 8K generation length (\(-\mathbf{0.3}\)\,pp) but a clear
advantage at 16K generation length (\(+\mathbf{1.8}\)\,pp). 
MMLU-Pro improves modestly but consistently (\(+\mathbf{0.4}\) and \(+\mathbf{1.1}\)\,pp) across both 8K and 16K.
These results suggest that expert diversity benefits harder multi-step scientific tasks most under longer generation budgets.
 
\item\textbf{Code generation.}
The largest gain appears on Codeforces at 16K generation length where the result is the most striking. \textsc{VP\(_2\)O} surpasses the baseline by \textbf{\(+\mathbf{179}\) ELO} and \(+\mathbf{3.6}\) Pass@1 points, while 8K shows near-parity. The gap emerges because longer generation windows allow essentially the Stein repulsive force to push
experts toward structurally distinct solution strategies, which is
particularly valuable in competitive programming where multiple correct
approaches exist.
 
\item\textbf{Instruction following.}
\textsc{VP\(_2\)O} delivers the largest and most consistent gains on
instruction following benchmarks. For IFBench loose and strict, specifically, the gains increased from roughly \(+\mathbf{3.6}\)  to \(+\mathbf{4.7}\)\,pp at 8K, widening to
\(+\mathbf{5.2}\) to \(+\mathbf{5.7}\)\,pp at 16K across all four metrics. This is consistent with the hypothesis
that expert specialization reduces reliance on a narrow set of high-reward response templates,
although routing diagnostics would be needed to confirm this mechanism directly. 
\end{itemize}
\begin{table}[ht]
\caption{%
  \textsc{VP\(_2\)O} vs. \textsc{GRPO}-baseline on a
  33B/4B-parameter \text{MoE}.  Results are averages over all evaluation
  checkpoints; $\Delta\!\uparrow$ is absolute improvement of
  \textsc{VP\(_2\)O} over the baseline (positive = better).
}
\label{tab:main_results}
\centering
\resizebox{\textwidth}{!}{%
\begin{tabular}{llccccccc}
\toprule
& & \multicolumn{3}{c}{\textbf{8K Context}} & \phantom{ab}
  & \multicolumn{3}{c}{\textbf{16K Context}} \\
\cmidrule(lr){3-5}\cmidrule(lr){7-9}
\textbf{Category} & \textbf{Benchmark}
  & Baseline & \textsc{VP\(_2\)O} & $\Delta\!\uparrow$
  &
  & Baseline & \textsc{VP\(_2\)O} & $\Delta\!\uparrow$ \\
\midrule
\multirow{2}{*}{\textit{Math}}
  & AIME 2024 (\%)
    & 68.2 & \textbf{70.8} & $+$2.6 &
    & 75.9 & \textbf{77.5} & $+$1.6 \\
  & AIME 2025 (\%)
    & 59.6 & \textbf{62.5} & $+$2.8 &
    & 67.8 & \textbf{69.9} & $+$2.1 \\
\midrule
\textit{Science}
  & GPQA Diamond (\%)
    & \textbf{64.2} & 63.9 & $-$0.3 &
    & 65.2 & \textbf{67.0} & $+$1.8 \\
\midrule
\textit{Knowledge}
  & MMLU-Pro (\%)
    & 71.1 & \textbf{71.5} & $+$0.4 &
    & 71.4 & \textbf{72.5} & $+$1.1 \\
\midrule
\multirow{2}{*}{\textit{Code}}
  & Codeforces ELO
    & \textbf{1419} & 1416 & $-$4 &
    & 1487 & \textbf{1666} & $+$179 \\
  & Codeforces Pass@1 (\%)
    & \textbf{24.4} & 24.3 & $-$0.1 &
    & 26.6 & \textbf{30.2} & $+$3.6 \\
\midrule
\multirow{4}{*}{\textit{Instruction}}
  & IFBench Loose Prompt (\%)
    & 63.7 & \textbf{68.3} & $+$4.6 &
    & 66.5 & \textbf{72.2} & $+$5.7 \\
  & IFBench Loose Instr.\ (\%)
    & 66.3 & \textbf{71.0} & $+$4.7 &
    & 69.3 & \textbf{74.9} & $+$5.6 \\
  & IFBench Strict Prompt (\%)
    & 58.1 & \textbf{61.7} & $+$3.6 &
    & 61.0 & \textbf{66.2} & $+$5.2 \\
  & IFBench Strict Instr.\ (\%)
    & 60.8 & \textbf{64.6} & $+$3.8 &
    & 63.7 & \textbf{69.0} & $+$5.3 \\
\bottomrule
\end{tabular}}
\end{table}
 
\subsection{Training Dynamics}
\label{sec:training_dynamics}
 
\begin{figure}[t]
  \centering
  \includegraphics[width=\textwidth]{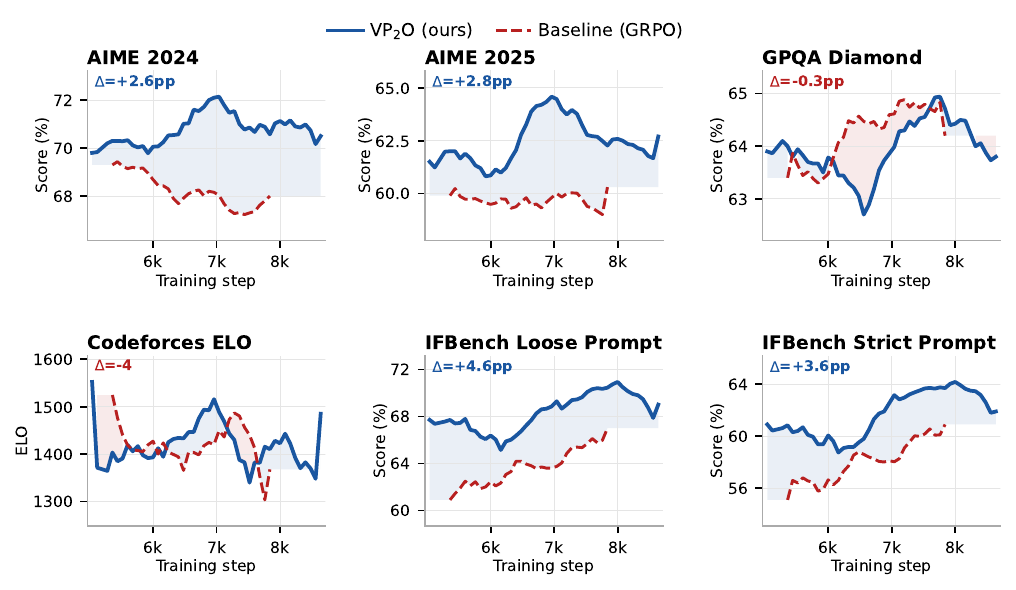}
  \caption{%
    Performance curves on academic benchmarks under the 8K
    generation context.  \textcolor{blue}{\textbf{Solid blue}}: \textsc{VP\(_2\)O}; \textcolor{red}{\textbf{dashed red}}:
    \textsc{GRPO} baseline.  Shaded region marks the leading method's
    advantage.  $\Delta$ values are checkpoint averages.
    \textsc{VP\(_2\)O} holds a consistent lead on AIME and IFBench
    throughout training; Codeforces ELO is near-parity at this context
    length (see Figure~\ref{fig:acc_16k} for the 16K result).
  }
  \label{fig:acc_8k}
\end{figure}

\begin{figure}[ht!]
  \centering
  \includegraphics[width=\textwidth]{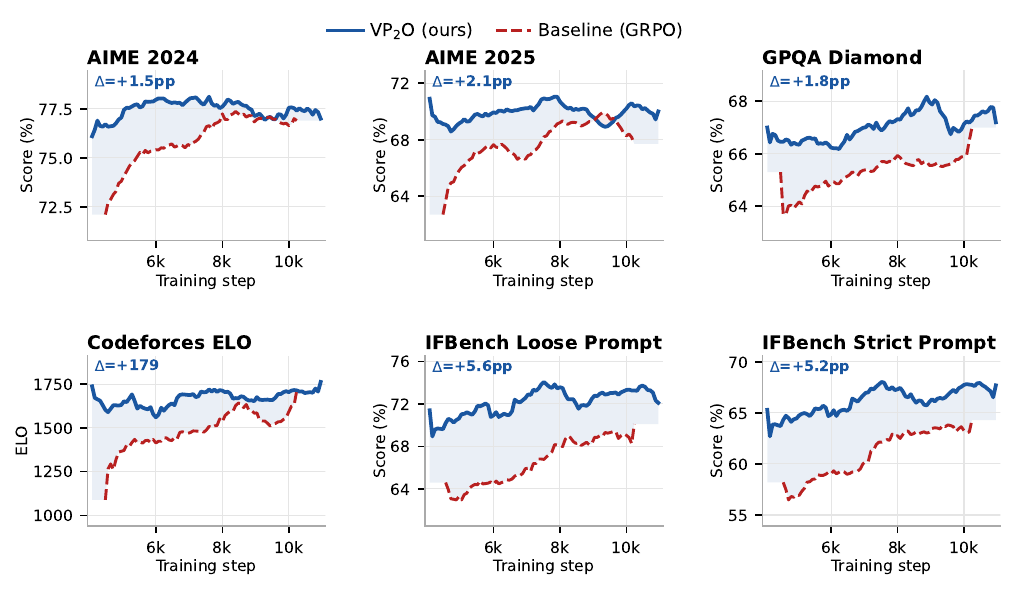}
  \caption{%
    Performance curves under the 16K generation length.
    Advantages are larger and more consistent than at 8K generation.
    Codeforces ELO shows the most striking result: \textsc{VP\(_2\)O}
    leads by $\mathbf{+179}$\,ELO from step \({\sim}4{,}000\) onward.
    On AIME 2024 the baseline converges toward \textsc{VP\(_2\)O}'s
    level, but around \(2{,}000\) steps later, confirming
    faster convergence rather than a vanishing advantage.
  }
  \label{fig:acc_16k}
\end{figure}

Figures~\ref{fig:acc_8k} and~\ref{fig:acc_16k} show training
trajectories for the six academic benchmarks.  Three patterns
stand out.
\begin{enumerate}[leftmargin=*]
\item\textbf{Stable advantage from early training.}  On every
benchmark where \textsc{VP\(_2\)O} leads, the gap is established
within the first few thousand steps and does not erode, consistent
with the geometric trust regions (Section~\ref{sec:sd}) providing
well-calibrated, data-driven step-size control from the start.
 
\item\textbf{Faster convergence at 16K.}  On AIME 2024, the baseline
requires roughly ${\sim}2{,}000$ additional steps to reach the
performance level \textsc{VP\(_2\)O} achieves at step ${\sim}4{,}000$
(see Figure~\ref{fig:acc_16k}, top left). Although, the baseline eventually approaches the
\textsc{VP\(_2\)O} performance level in late training, the reduced training steps suggests a convergence-
speed benefit rather than only a final-score benefit.
 
\item\textbf{Baseline degradation at 8K.}  On AIME 2024 and Codeforces the baseline shows 
a declining trend in late training while \textsc{VP\(_2\)O} remains stable. This behavior 
is consistent with reward over-optimization or reward-mode collapse, two forms of reward 
hacking that the Stein repulsive force can help mitigate. A stronger diagnosis would require, 
however, additional measurements such as router entropy, expert utilization, expert-overlap scores, and reward-model score trajectories.
\end{enumerate}
 
\subsection{Solution Token Efficiency}
\label{sec:token_efficiency}
 
\begin{figure}[t]
  \centering
  \includegraphics[width=\textwidth]{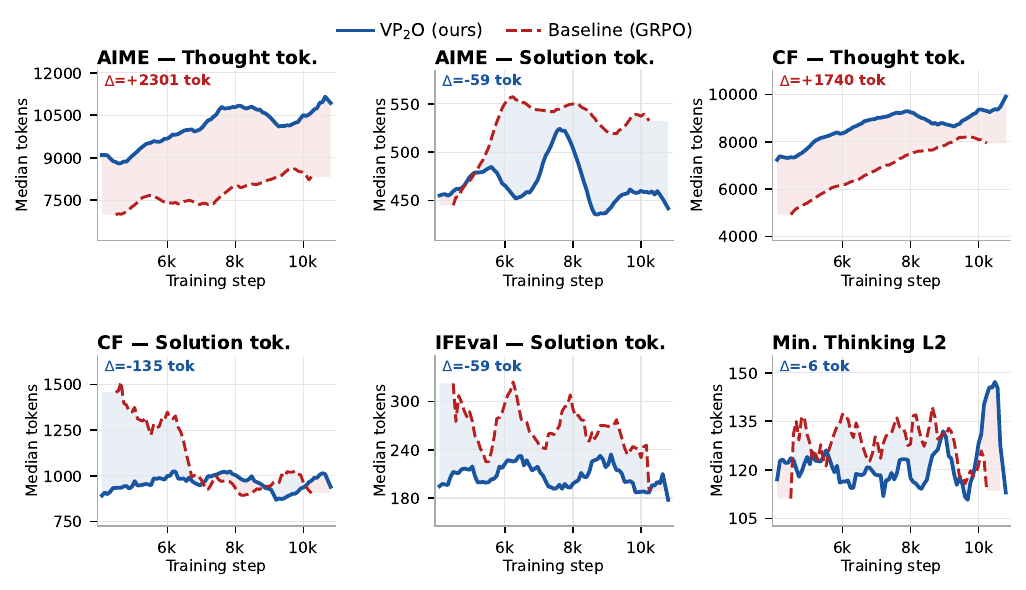}
  \caption{%
    Thought and solution token counts per task throughout training
    (\textbf{16K} context). \textcolor{blue}{Blue shading}: \textsc{VP\(_2\)O}
    requires fewer tokens; \textcolor{red}{red shading}: baseline
    requires fewer tokens.  \textsc{VP\(_2\)O} uses more \emph{thought}
    tokens on AIME (\(+\mathbf{2,301}\)) and Codeforces (\(+\mathbf{1,740}\)) while emitting
    fewer \emph{solution} tokens, reflecting deeper but more concise
    reasoning.  IFEval solution tokens fall by 59 tokens (\(-\mathbf{22}\)\%).
  }
  \label{fig:tok_16k}
\end{figure}

Beyond accuracy, we examine whether \textsc{VP\(_2\)O} learns more
\emph{efficient} solution strategies, measured by median solution
token count.  Table~\ref{tab:tokens} summarizes the results;
Figure~\ref{fig:tok_16k} plots the training trajectories.
\begin{table}[t]
\caption{%
  Median solution token counts.  Fewer tokens at equal or higher
  accuracy indicates more efficient reasoning.
  $\Delta$ is \textsc{VP\(_2\)O} minus Baseline; \textbf{bold}
  marks the more efficient model.
}
\label{tab:tokens}
\centering
\begin{tabular}{lcccccc}
\toprule
& \multicolumn{3}{c}{\textbf{8K Context}}
& \multicolumn{3}{c}{\textbf{16K Context}} \\
\cmidrule(lr){2-4}\cmidrule(lr){5-7}
\textbf{Task} & Base & \textsc{VP\(_2\)O} & $\Delta$
              & Base & \textsc{VP\(_2\)O} & $\Delta$ \\
\midrule
AIME 2025        & 409 & \textbf{273} & $-$130 & 527 & \textbf{468} & $-$59  \\
Codeforces       & \textbf{791} & 1176 & $+$376 & 1096 & \textbf{961} & $-$135 \\
IFEval           & 213 & 232          & $+$19  & 267  & \textbf{208} & $-$59  \\
Min.\ Thinking L1 & 158 & \textbf{138} & $-$20 & 159  & \textbf{155} & $-$5   \\
Min.\ Thinking L2 & 159 & \textbf{102} & $-$57 & 129  & \textbf{123} & $-$6   \\
\bottomrule
\end{tabular}
\end{table}
 
\textbf{AIME.}
\textsc{VP\(_2\)O} uses \textbf{130 fewer solution tokens} (\(-\mathbf{32}\)\%)
at 8K and 59 fewer (\(-\mathbf{11}\)\%) at 16K while achieving higher accuracy.
The thought-token trajectory (Figure~\ref{fig:tok_16k}) reveals the
complementary picture: \textsc{VP\(_2\)O} ``thinks" \emph{more} during
reasoning but writes \emph{less} in
its final answer. This suggests a shift toward longer intermediate reasoning and shorter final answers, although trace-level analysis would be needed to determine whether the additional thought tokens correspond to better reasoning or not.
 
\textbf{Codeforces.}
On coding, the pattern depends on generation length. At 8K, \textsc{VP\(_2\)O} produces longer code solutions (\(+\mathbf{376}\)
tokens) with near identical \textsc{ELO}, suggesting no clear efficiency gain in that setting. At 16K, however, \textsc{VP\(_2\)O} produces shorter solutions while substantially improving \textsc{ELO} and Pass@1,
making this the clearest case where quality and solution-token efficiency improve together.
 
\textbf{Instruction following.}
On IFEval, \textsc{VP\(_2\)O} reduces solution length at 16K by 59 tokens. This outcome is
consistent with the model satisfying instructions more precisely, but the result should be interpreted as
output-token efficiency, rather than improved reasoning efficiency where more evidence is needed.

\subsection{Discussion}
 
\textbf{When does VP\(_2\)O help most?}
\textsc{VP\(_2\)O} appears most helpful in three regimes: longer generation
budgets, tasks that admit multiple solution strategies, and early-to-mid training phases where expert
diversity has not yet collapsed. The strongest evidence appears at 16K on Codeforces and instruction
following, where \textsc{VP\(_2\)O} improves both quality and solution-token efficiency. GPQA
at 8K is the main exception: \textsc{VP\(_2\)O} trails by \(\mathbf{0.3}\) pp, but the 16K result recovers to a \(+\mathbf{1.8}\) pp
advantage.
 
\textbf{Computational overhead.}
The Stein transport field \(\varphi(\theta_i)\) requires one
additional forward pass per expert to compute the prototype update
and co-activation scores, adding approximately 5--8\% wall-clock
overhead relative to our \textsc{GRPO} baseline in this implementation. The geometric trust-region check is \(\mathcal{O}(N D_{\text{out}})\) per step and is
negligible relative to the model forward/backward pass.
 
\textbf{Limitations.}
Our evaluation is limited to a single model family and scale.
Whether the variational benefits persist at larger scales (e.g. 70B+)
or with more experts (\(N\gg20\)) is an important open question. The current work also does not report on routing diagnostics such as router entropy, expert utilization, expert-overlap measures, KL drift, or ESS trajectories. 
Future work should test whether the same benefits hold across different model families, expert counts, and reward/data mixtures.

\section{Conclusion}
We presented \textsc{VP\(_2\)O}, a variational proximal policy optimization framework that views regularized alignment optimization, such as \textsc{PPO/GRPO}, as particle-based variational inference within sparse Mixture-of-Experts architectures. By using functional kernels defined over low-dimensional expert prototypes, \textsc{VP\(_2\)O} introduces geometry-aware proximal controls that complement or partially replace fixed clipping and KL divergence schedules. Empirical evaluations on a single 33B/4B sparse \text{MoE} model show consistent improvements on several mathematical reasoning, code-generation, and instruction-following benchmarks, with the strongest gains appearing at longer generation lengths. These results suggest that preserving functional diversity across expert parameters is a promising direction for stabilizing long-form reinforcement-learning trajectories in large language models, while broader validation across model families, scales, reward and data mixtures remains future work.

\Urlmuskip=0mu plus 1mu\relax
\bibliographystyle{plainnat}  
\bibliography{refs}


\appendix




\newpage

\end{document}